\documentclass[conference]{IEEEconf}
\IEEEoverridecommandlockouts
\usepackage{moreverb,url}
\usepackage[T1]{fontenc}
\usepackage{float}
\usepackage{wrapfig}
\usepackage{amsfonts}
\usepackage{enumerate}
\usepackage[linesnumbered,ruled]{algorithm2e}
\usepackage{amsthm}
\usepackage{pifont}
\usepackage{xcolor}
\usepackage{soul}

\newtheorem{definition}{Definition}
\newtheorem{proposition}{Proposition}

\newtheorem{problem}{Problem}
\usepackage{listings}
\usepackage{amsmath}
\usepackage{amssymb}
\usepackage{subfig}
\usepackage{graphicx}
\usepackage{csquotes}

\newcommand{\deleted}[1]{}

\renewcommand{\baselinestretch}{0.943}
\begin{document}
\title{\bf 
Finding Optimal Modular Robots for Aerial Tasks
}
\author{Jiawei Xu and
David Salda\~na
\thanks{J. Xu and D. Salda\~{n}a are with the Autonomous and Intelligent Robotics Laboratory (AIRLab), Lehigh University, PA, USA:$\{$\texttt{jix519, saldana\}@lehigh.edu}}
}

\maketitle
\begin{abstract}
Traditional aerial vehicles have limitations in their capabilities due to actuator constraints, such as motor saturation. The hardware components and their arrangement are designed to satisfy specific requirements and are difficult to modify during operation. To address this problem, we introduce a versatile modular multi-rotor vehicle that can change its capabilities by reconfiguration. Our modular robot consists of homogeneous cuboid modules, propelled by quadrotors with tilted rotors. Depending on the number of modules and their configuration, the robot can expand its actuation capabilities. In this paper, we build a mathematical model for the actuation capability of a modular multi-rotor vehicle and develop methods to determine if a vehicle is capable of satisfying a task requirement. Based on this result, we find the optimal configurations for a given task. Our approach is validated in realistic 3D simulations, showing that our modular system can adapt to tasks with varying requirements.

\end{abstract}

\section{Introduction}{
The last decade witnessed the development of multi-rotor vehicles which
use thrust forces from the propellers to compensate for gravity and to maneuver. 
Dynamic models have been proposed in the literature to describe and control the motion of multi-rotor vehicles~\cite{1302409,7139759,alaimo2013mathematical,beniak2016control}.
Despite their numerous advantages, the lack of versatility remains a significant concern for multi-rotor vehicles. For instance, a light-weight quadrotor carrying a camera has limited potential to increase the payload and transport heavy objects. 
Furthermore, multi-rotor vehicles with vertical rotors are underactuated and require tilting for horizontal translation.
These limitations in strength and actuation arise when the vehicle is designed specifically for certain tasks, thus limiting their versatility.

To address the lack of versatility in multi-rotor designs, researchers have present various approaches.
For example, additional vertically-placed rotors in a multi-rotor can increase the strength of the vehicle, enabling it to transport heavier objects~\cite{Oung2011TheArray,alaimo2013mathematical}. Modular vehicles based on quadrotors have also been developed, where the quadrotors rigidly attach to each other to increase strength and form different shapes on the fly~\cite{8461014}. However, such aerial systems do not increase their actuated degrees of freedom (ADOF) as long as the rotors share the same orientation.

\begin{figure}
\captionsetup[subfigure]{labelformat=empty}
    \centering
    \subfloat[]{\includegraphics[width=0.5\linewidth]{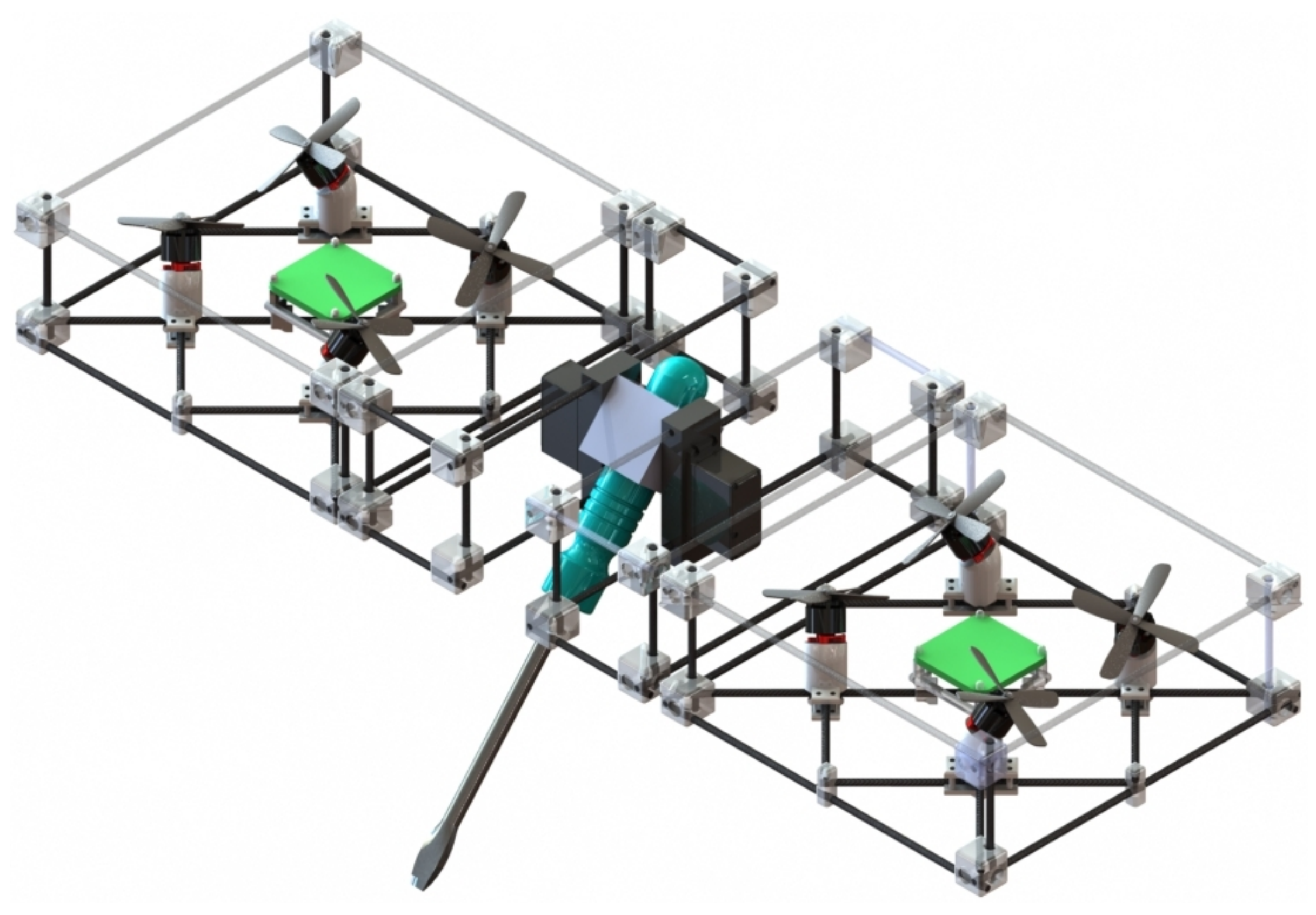}}
    \subfloat[]{\includegraphics[width=0.5\linewidth]{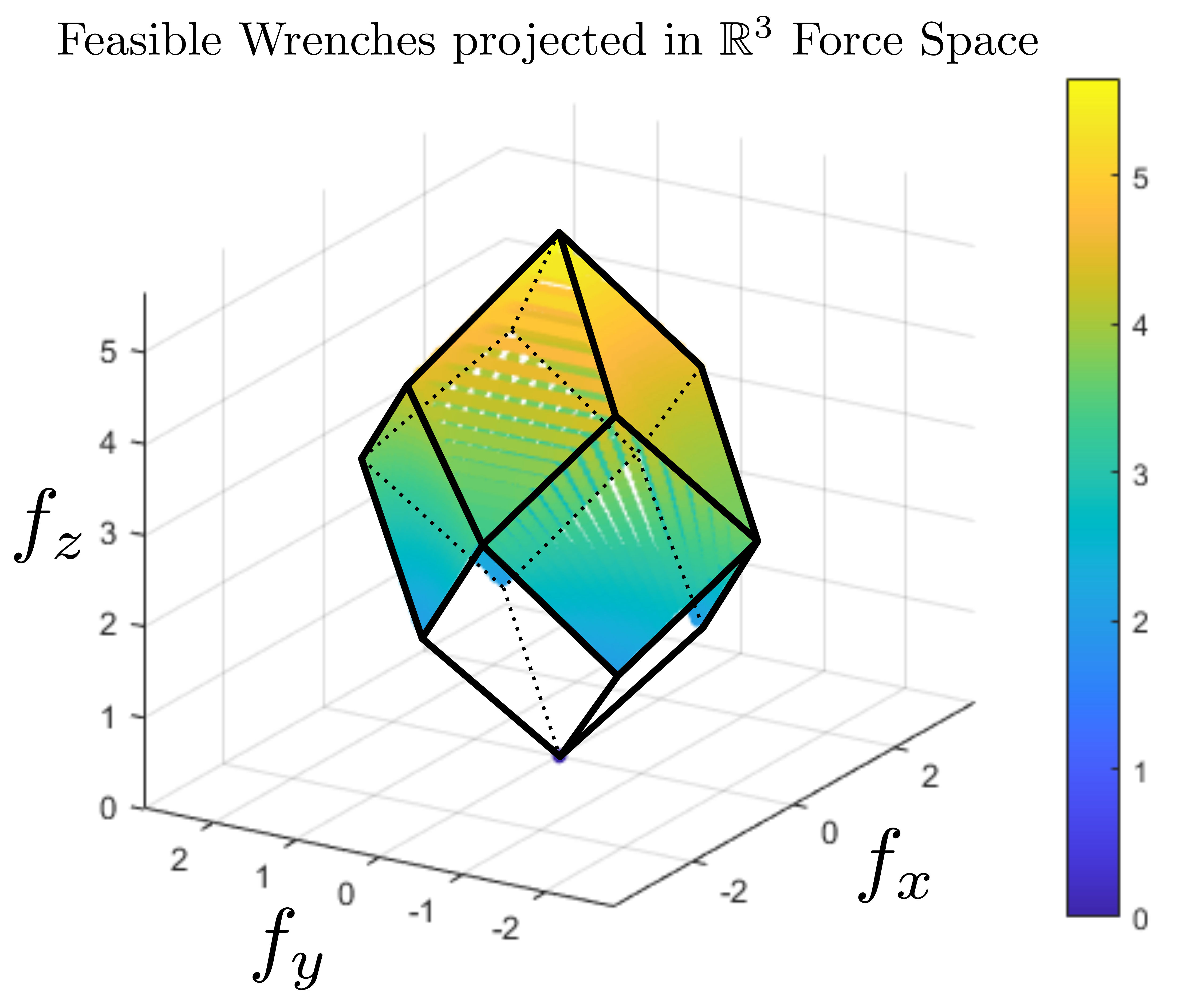}}
    \vspace{-1em}
    \caption{A modular multi-rotor manipulating a screwdriver. The set of all forces that the vehicle can generate is represented by a polytope
    in the $\mathbb{R}^3$ force space.}
    \label{fig:manipulation}
    \vspace{-1em}
\end{figure}

Incorporating modularity in multi-rotor vehicles is a scalable approach to improve their strength and actuation. In~\cite{8258850}, the authors present a robot composed of modules actuated by pairs of actively tilting rotors. Our previous work presents H-ModQuad~\cite{jiawei2021HModQuad}, a heterogeneous modular system composed of modules that fly autonomously and change their ADOF by reconfiguration. These approaches improve the versatility of multi-rotor vehicles by expanding their actuation capabilities, \textit{i.e.,} the maximum force and torque they can generate. 
In a modular system, it is essential to determine whether there exists a configuration that satisfies the task requirements or not. If so, we need to find an optimal configuration that minimizes the number of modules.

In~\cite{9636086}, the authors present a modular flying robot that can reconfigure by tilting the modules inside the structure to adapt to different trajectories; to determine whether a configuration satisfies a task requirement or not, the vehicle evaluates several configurations to track a trajectory in simulation. 
We propose a geometric approach to solve this problem by modeling the task requirements and the actuation capabilities of a multi-rotor vehicle using polytopes.
In recent literature, we notice methods to describe the actuation capability of multi-rotor vehicles~\cite{7759271,hamandi2021design,6161298} that are commonly illustrated as a polytope within which forces and torques are achievable by the vehicles. 
Researchers also present optimizations for the design of multi-rotor vehicles with performance characteristics, both in theory~\cite{9438375} for the rotor orientation, and in practice~\cite{6935598} for the hardware components. These methods work well for either fixed designs or fixed tasks such as following a trajectory. In this paper, we develop a method that allows to check whether a modular multi-rotor vehicle satisfies a task requirement or not. We reconfigure the vehicle by leveraging on modularity to expand its capabilities and satisfy the task requirements. Fig.~\ref{fig:manipulation} illustrates a potential application of our method where we can find an appropriate multi-rotor configuration to operate a screwdriver.
We consider the reconfiguration as an offline procedure which is not affected by any physical constraints such as the space limit~\cite{halperin2020space}.

The main contribution of this paper is threefold. First, we provide a formal definition of aerial task requirements present a geometric model that characterizes the feasible wrenches of a multi-rotor vehicle with uni-directional rotors. Second, we present a convex hull abstraction of the actuation capabilities of a multi-rotor vehicle,
and an optimization-based algorithm to identify whether a configuration satisfies a set of task requirements. Third, we develop search based method that makes a modular multi-rotor vehicle expand its capabilities to satisfy the task requirements. Although we analyze the methods with H-ModQuad modules, the abstractions and algorithms are applicable to any modular multi-rotor vehicles.
}

\section{Background on Modular Multi-Rotors}
\label{sec:bg}
Integrating modularity into multi-rotor systems enhances their versatility. Modular systems can adapt to different tasks through self-reconfiguration. This concept has sparked a wide range of discussion, including the fractal assembly~\cite{9172614}, cooperative transport~\cite{Mellinger2012CooperativeQuadrotors,9364354}, novel structural design~\cite{duffy2015lift,jiawei2021HModQuad}, and various applications~\cite{8461014,modquadgripper}.

In~\cite{8461014}, the authors present a modular aerial multi-rotor vehicle design called ModQuad, where homogeneous quadrotor modules assemble in mid-air to compose a structure. Our previous work presents H-ModQuad~\cite{jiawei2021HModQuad}, which integrates heterogeneous quadrotor modules with tilted rotors in different directions. This change allows the vehicle to increase the number of ADOF. This paper focuses on modular aerial multi-rotors composed of homogeneous modules with tilted motors. 
\begin{definition}[Module]
A \emph{module} is an aerial vehicle composed of a quadrotor within a cuboid frame.
The propellers do not have to be vertical with respect to the base of the module, and their orientation determines the actuation properties of the module. 
\end{definition}

\noindent
In contrast to our previous work on heterogeneous modules, this work is focused on using a single type of module to assemble a versatile robot.
Based on an attaching mechanism on the frame, 
a pair of modules can dock and create a rigid connection by aligning their vertical faces.
Via docking actions, we assemble a \emph{structure}.

\begin{definition}[Structure]
A \emph{structure} is a group of $n\geq1$ rigidly-connected modules docked horizontally that function as a single multi-rotor vehicle.
\end{definition}

\begin{figure}[t]
    \centering
    \vspace{0.5em}
    \includegraphics[width=0.7\linewidth]{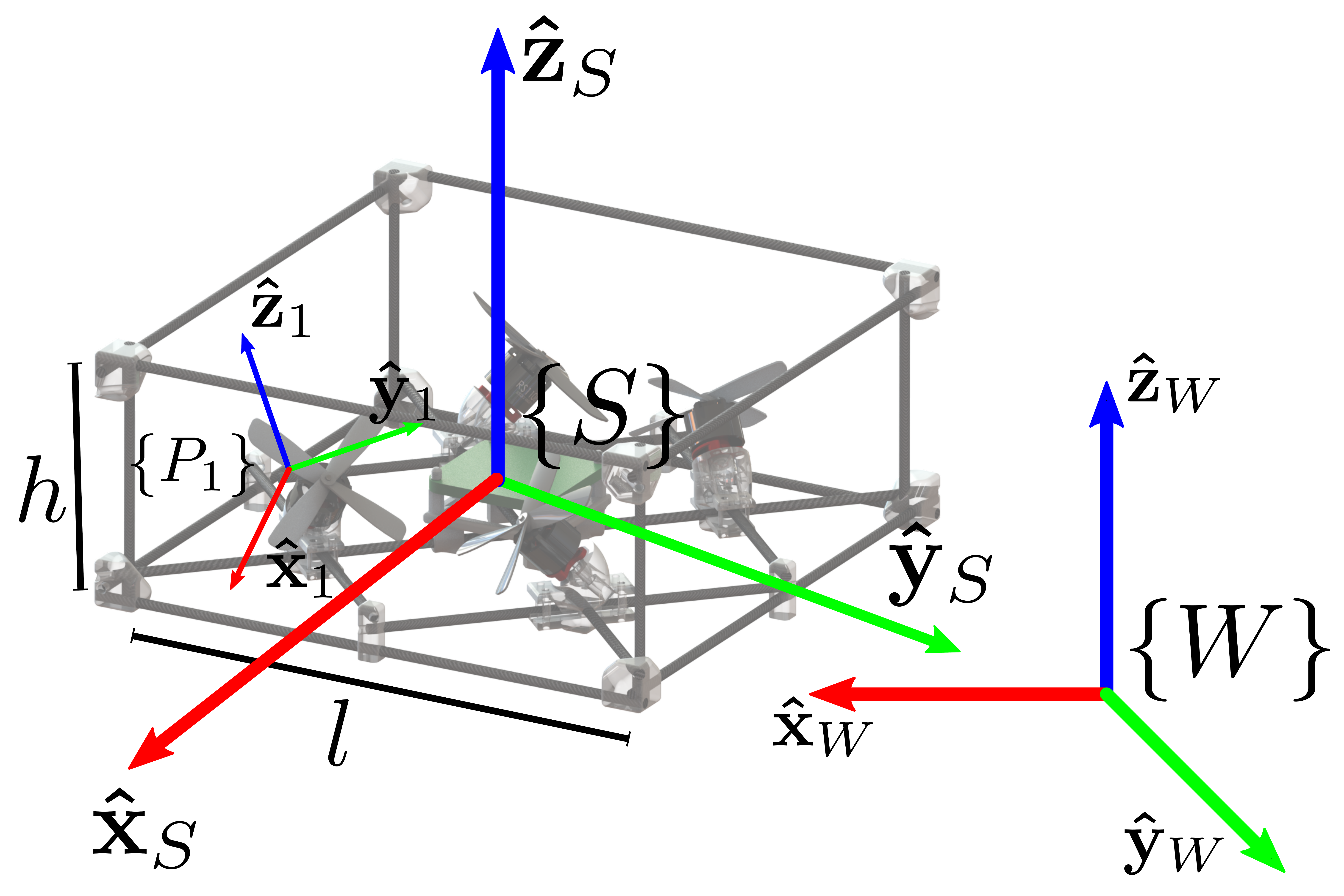}    
    \caption{A module with its coordinate frames and dimensions labelled. $\{P_1\}, \{S\}, \{W\}$ are the frames of the $1^\text{st}$ rotor, the structure, and the world reference, respectively. Note that this module composes a structure, thus $\{S\}$ aligns with $\{M_1\}$.}
    \label{fig:oneUAV}
    \vspace{-1em}
\end{figure}

\noindent
We denote the unit basis in $\mathbb{R}^3$ by~$\boldsymbol{e}_1=[1,0,0]^\top,\: \boldsymbol{ e}_2=[0,1,0]^\top,$ and $\boldsymbol{ e}_3=[0,0,1]^\top$. 
The $i$-th module in the structure has a module reference frame, $\{M_i\}$, with its origin in the module's center of mass (COM). 
The four propellers are in a square configuration and located on the $xy$-plane. 
A structure has $4n$ rotors. For each rotor $k\in\{1, 2, \dots, 4n\}$, we define a propeller frame $\{P_{k}\}$ with its $z$-axis pointing in the direction of the propeller force and its $x$-axis points towards the front of the module. The coordinate frames of a module are illustrated in Fig. \ref{fig:oneUAV}, where $\{W\}$ stands for the world reference frame. The structure frame, denoted by~$\{S\}$, has its origin in its COM. Without loss of generality, we align all the module frames in the structure and define the $x$-, $y$- and $z$-axes of $\{S\}$ as in parallel to the $x$-, $y$- and $z$-axes of all modules in the structure. 
%

Once a configuration is assembled, the position and orientation of all rotors are fixed with respect to $\{S\}$. We assume all $4n$ rotors on the structure are identical, and each of them can generate a thrust force $f_k$ and a torque from the air drag. The actuation of the rotors generates a wrench that applies on the structure, which is described by a linear mapping,
$\boldsymbol{w} = \boldsymbol{A}\boldsymbol{u}$,
where the input $\boldsymbol{u}=[f_1, f_2, \dots, f_{4n}]^\top$ is the thrust force of all rotors, and the wrench vector
$\boldsymbol{w} = \left[\boldsymbol{f}^\top\; \boldsymbol\tau^\top\right]^\top$
is a $6\times1$ vector of force and torque vectors. 
The \emph{configuration matrix}, 
\begin{equation}
    \centering
    \boldsymbol{A} = \begin{bmatrix}
    \cdots & {}^{S}\!\boldsymbol{R}_{k}\boldsymbol{e}_3 & \cdots \\
    \cdots & \boldsymbol{p}_{k}\times{}^S\!\boldsymbol{R}_{k}\boldsymbol{ e}_3+{}^{S}\!\boldsymbol{R}_{k}(-1)^{1+k}c_\tau\boldsymbol{ e}_3 & \cdots
    \end{bmatrix},
    \label{eq:Acomponents}
\end{equation}
is a $6\times4n$ matrix that maps the input forces into the total wrench in $\{S\}$, where ${}^{S}\boldsymbol{R}_{k}\in\mathsf{SO}(3)$ and $\boldsymbol{p}_{k}\in\mathbb{R}^3$ describe the orientation and position of the $k$-th rotor with respect to $\{S\}$, respectively. $c_\tau$ is a coefficient related to the motors and propellers that can be obtained experimentally. 
In traditional multi-rotor vehicles, for instance, a quadrotor, the matrix $\boldsymbol{A}$ has dimension $6\times4$ and is fixed after building the robot~\cite{6225129,5569026}. In our case, 
the dimension of the matrix can grow in groups of four columns by increasing the number of modules, and its values will depend on the structure configuration.
The configuration matrix describes the actuation properties of a structure. For instance, the number of ADOF is equal to the rank of the configuration matrix~\cite{jiawei2021HModQuad}.

\section{Problem statement}
\label{sec:problem}
The configuration of a multi-rotor structure dictates how the input forces are mapped to the total wrench, which depends on the \emph{rotor configuration}, \textit{i.e.,} the rotors' position and orientation of the multi-rotor structure.
\begin{definition}[Rotor Configuration]
The \emph{rotor configuration} of a structure, $\mathcal{D}$, is defined as a set of tuples $\mathcal{D}=\{(\boldsymbol{p}_{k}, {}^{S}\!\boldsymbol{R}_{k})\vert k = 1, 2, \dots, 4n\}$ that represents each rotor's position and orientation with respect to the structure frame.
\end{definition}
\noindent
The actuation capability of a multi-rotor vehicle is defined as the set of all \emph{feasible wrenches} based on its configuration matrix $\boldsymbol{A}$ and the actuation limit of the input force, $f_{max}$. 
\begin{definition}[Feasible Wrenches]
    The set of feasible wrenches, $\mathcal{W}$, defines all available wrenches $\mathcal{W}$ that a structure with a configuration matrix $\boldsymbol{A}$ can generate, \textit{i.e.,}
    \begin{equation}
    \mathcal{W} = \left\{\boldsymbol{A}\boldsymbol{u}\:\vert\: 0\preceq \boldsymbol{u}\preceq f_{max}, \boldsymbol{u}\in\mathbb{R}^{4n}\right\},
    \label{eq:volume}
    \end{equation}
\end{definition}

\noindent
where the operator ``$\preceq$'' represents element-wise comparison. Considering a structure with uni-directional motors, all propellers are identical, and they can produce continuous non-negative thrust force which has a upper bound $f_{max}$.
Aerial tasks involve physical interaction with the environment and require to generate certain wrenches. 

\begin{definition}[Task Requirement]
    A \emph{task requirement} is a set of wrenches $\mathcal{T}\subset\mathbb{R}^6$ that a robot needs to generate to perform a task.
\end{definition}

\noindent
A structure with feasible wrenches $\mathcal{W}$ satisfies a task requirement $\mathcal{T}$, if $\mathcal{T}\subseteq\mathcal{W}$. For instance, to manipulate a screwdriver, a multi-rotor needs to generate both a torque for rotation and a force for translating the screwdriver. Thus, the task requirement is $\mathcal{T} = {\boldsymbol{w}_1, \boldsymbol{w}_2}$, where $\boldsymbol{w}_1$ is the wrench for translation and alignment, and $\boldsymbol{w}_2$ is for driving the screw. Our objective is to determine whether a given structure satisfies a task requirement, and if not, to enhance the structure capabilities by adding modules until it satisfies the requirement. We formally define the problem as follows.

\begin{problem}[Finding structure configurations]
Given a task requirement $\mathcal{T}$, find the structure configuration $\mathcal{D}$ with the minimum number of modules that is able to satisfy the task requirement.
\label{p:satisfyingtask}
\end{problem}
\noindent

\section{Structure-Task Requirement Evaluation}{
\label{sec:polytope}
The main challenge to solve Problem \ref{p:satisfyingtask} is that the set $\mathcal{W}$ has infinite elements, making it computationally difficult to explore. Notice that the constraints on the input vector $0\preceq\boldsymbol{u}\preceq f_{max}$ represent the intersection of $2\cdot4n$ half-spaces in the $\mathbb{R}^{4n}$ input space, which forms a convex polyhedron~\cite{boyd2004convex}. Since the element-wise comparison provides both upper and lower bounds for the polyhedron along each axis, the polyhedron of constraints is a bounded convex polytope~\cite{borrelli2017predictive}. The feasible wrenches are obtained by mapping the polytope with an affine function $\boldsymbol{A}$ to the $\mathbb{R}^6$ wrench space. Thus, $\mathcal{W}$ is a convex polytope in $\mathbb{R}^6$.
%

A closer inspection on \eqref{eq:volume} shows that
$\mathcal{W}$ is the Minkowski sum of the column vectors of $\boldsymbol{A}$ whose size is scaled by $f_{max}$, which means $\mathcal{W}$ is a zonotope. Such a zonotope is defined as a set in $\mathbb{R}^n$ constructed from generator vectors $\boldsymbol{g}_i$, $i = 1, \dots, 4n$, of which an element can be expressed as $\sum_i a_i\boldsymbol{g}_i, 0\leq a_i\leq 1$. Different choices of $a_i$ result in different elements, and the zonotope is the union of all such elements~\cite{fukuda2004zonotope,kuhn1998rigorously,coxeter1973regular,beck2007computing,grunbaum1967convex}. The generator vectors of the feasible wrenches $\mathcal{W}$ are the column vectors of $\boldsymbol{A}$. 
Based on this insight,
we provide two methods to determine whether the condition $\mathcal{T}\subseteq\mathcal{W}$ is satisfied or not. 

\subsection{Convex Hull Generation for Zonotopes}
\label{sec:hullG}
\begin{figure}[t]
\vspace{0.5em}
\centering
    \subfloat[A 2x2 structure is composed of four modules. The red arrows represent the direction of the rotors.\label{fig:examplestructure}]{\includegraphics[width=0.48\linewidth]{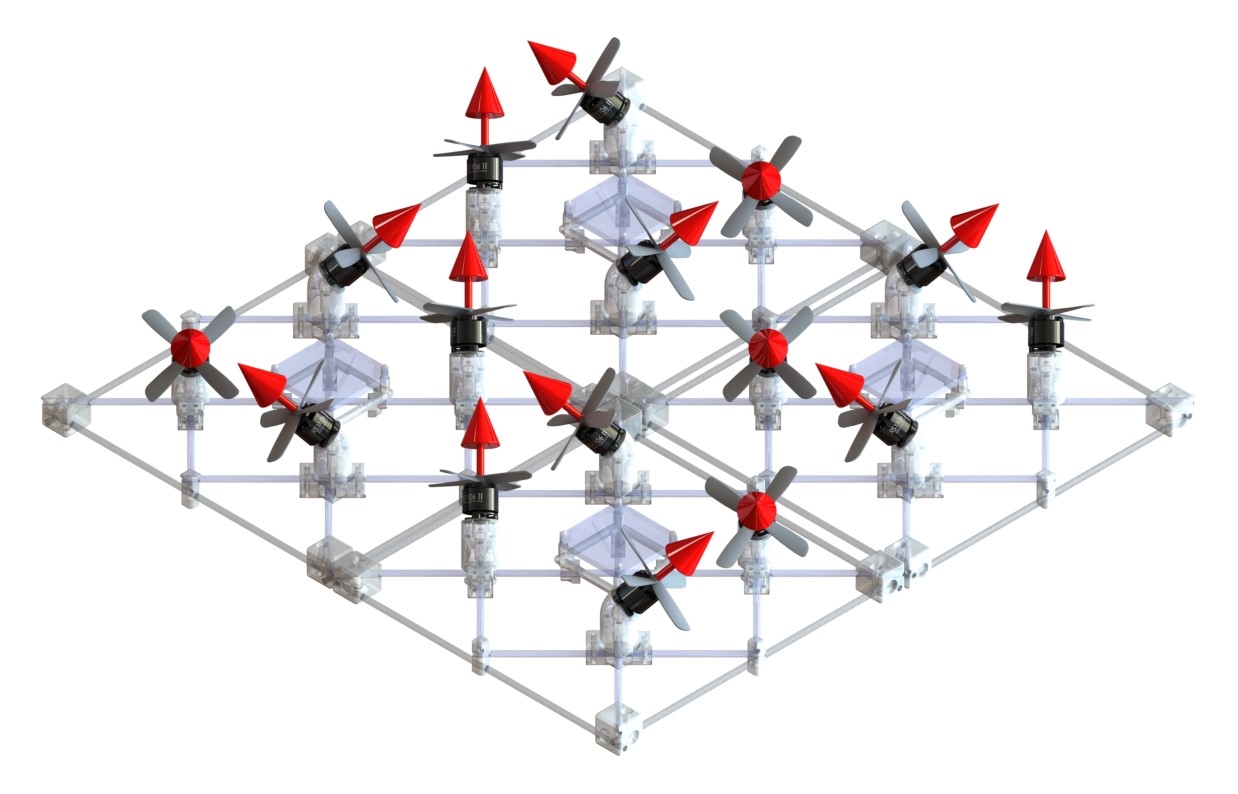}}
    \hspace{5pt}
    \subfloat[The polytope of the feasible wrenches $\mathcal{W}$ of the structure, projected in the torque space. \label{fig:polytopes}]{\includegraphics[width=0.48\linewidth]{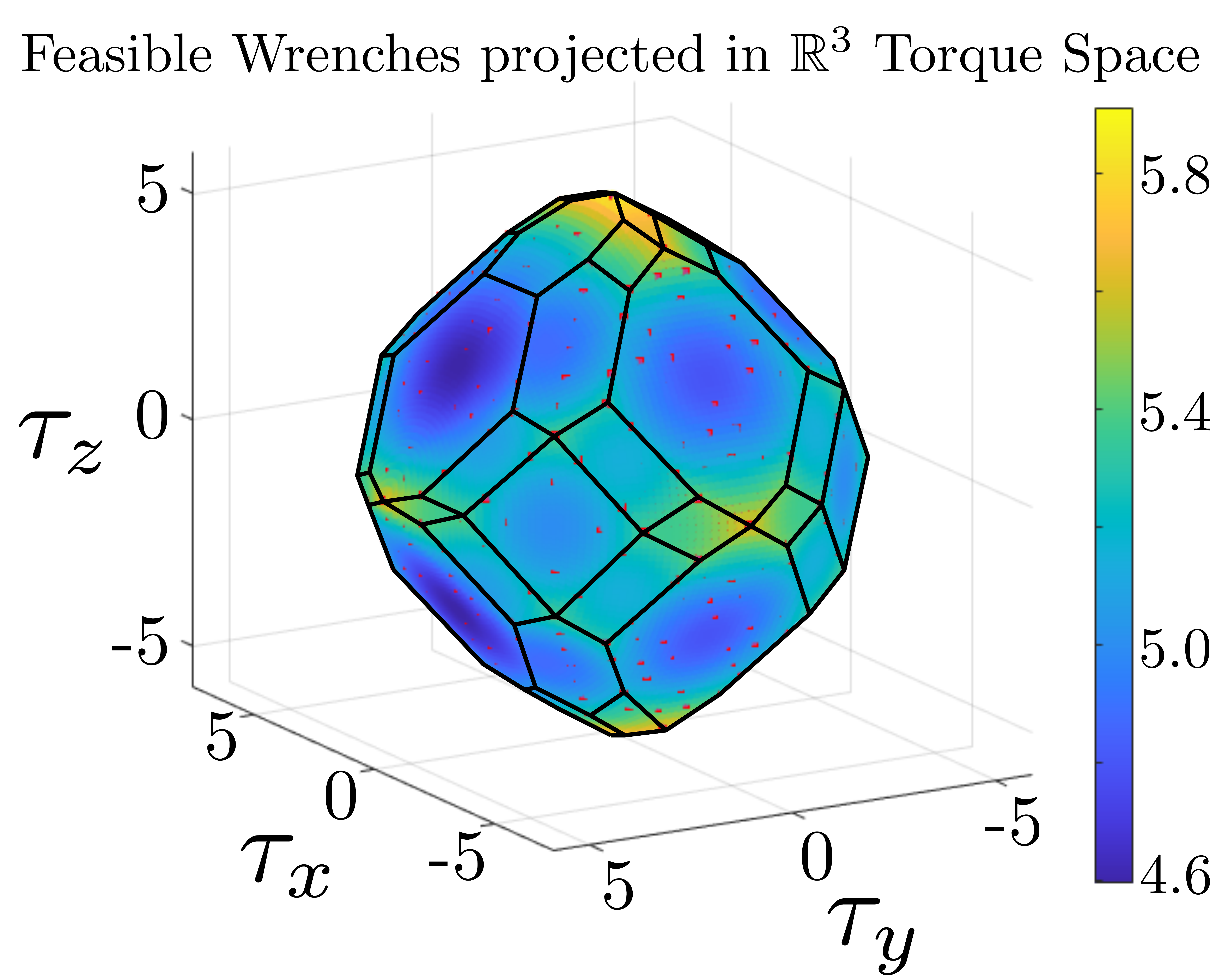}}
    \label{fig:structureandpolytope}
    \caption{A structure composed of four modules and a visualization of its feasible wrenches projected in the torque space.}
\end{figure}

\noindent
Since a zonotope is a polytope and the convex hull of a convex polytope is the polytope itself, 
we characterize the feasible wrenches $\mathcal{W}$ with the convex hull of $\mathcal{W}$, denoted as $\mathcal{H(W)}$. Furthermore, the convenience of the representation of a zonotope with only its generators allows us to generate $\mathcal{H(W)}$ directly from $\boldsymbol{A}$.
Thereby, one approach based on the concept of convex hull is to find $\mathcal{H(W)}$ using $\boldsymbol{A}$ and then check whether all required wrenches belong to it or not.

We start by introducing a discretization approach to construct the convex hull of the zonotope. 
First, discretize the feasible wrenches for binary input vectors which contain either $f_{max}$ or $0$, and group them in a set 
\begin{equation}
    \mathcal{S} = \{\boldsymbol{Au}\:\vert\: \boldsymbol{u}=\{0, f_{max}\}^{4n}\}.
    \label{eq:setS}
\end{equation}
Then, find the convex hull
\begin{equation}
    \mathcal{H(S)} = \{\Sigma_i\lambda_i\boldsymbol{s}_i\:\vert\:\Sigma_i\lambda_i=1, \lambda_i\geq0, \boldsymbol{s}_i\in\mathcal{S}\}
    \label{eq:HullS}
\end{equation}
of the set of points $\mathcal{S}$ using existing convex hull generation algorithms such as the Quickhull algorithm~\cite{barber1996quickhull}.
\begin{proposition}
The convex hull of the set $\mathcal{S}$, $\mathcal{H(S)}$, is equivalent to $\mathcal{W}$, if $\mathcal{W}$ and $\mathcal{S}$ are described by \eqref{eq:volume} and \eqref{eq:setS}, respectively.
\end{proposition}

\begin{proof}
To prove the equivalency, we show $\mathcal{H(S)}\subseteq\mathcal{W}$, then $\mathcal{W}\subseteq\mathcal{H(S)}$. First, by definition of the convex hull, 
for any arbitrary point $\boldsymbol{v}\in\mathcal{H(S)}$, 
there exists a vector $\boldsymbol{\lambda}=[\lambda_1,\dots, \lambda_{4n}]^\top$ such that $\boldsymbol{v}=\sum_i\lambda_i\boldsymbol{s}_i$ for all $\boldsymbol{s}_i\in\mathcal{S}$ where  $\lambda_i\geq0,$ and $\sum_i\lambda_i=1$. By definition of $\mathcal{S}$, any element in $\mathcal{S}$ can be expressed as $\boldsymbol{Au}$ for some $\boldsymbol{u}\in\{0, f_{max}\}^{4n}$. Substituting all $\boldsymbol{s}_i$ with $\boldsymbol{Au}_i$, we obtain $\boldsymbol{v}=\sum_i\lambda_i\boldsymbol{s}_i=\sum_i\lambda_i(\boldsymbol{Au}_i) = \boldsymbol{A}\sum_i(\lambda_i\boldsymbol{u}_i)$ for any $\boldsymbol{v}\in\mathcal{H(S)}$.
Let us define a vector $\boldsymbol{\sigma}\in[0,f_{max}]^{4n}$. 
We can express $\sum_i\lambda_i\boldsymbol{u}_i = \boldsymbol{\sigma}, 0\preceq\boldsymbol{\sigma}\preceq f_{max}$ 
because $\sum_i\lambda_i=1$ and $\boldsymbol{u}_i\in\{0, f_{max}\}$. Therefore, for any arbitrary $\boldsymbol{v}\in\mathcal{H(S)}$, there always exists $\boldsymbol{u}$ such that $\boldsymbol{v=Au}, 0\preceq\boldsymbol{u}\preceq f_{max}$, meaning that $\boldsymbol{v}\in\mathcal{W}$,
which shows $\mathcal{H(S)}\subseteq\mathcal{W}$. 

Second, we show $\mathcal{W}\subseteq\mathcal{H(S)}$. For an arbitrary point~$\boldsymbol{w}\in\mathcal{W}$, by the definition of $\mathcal{W}$, there exists an input vector~$\boldsymbol{u}$ such that $\boldsymbol{w=Au}, 0\preceq\boldsymbol{u}\preceq f_{max}$. 
The following procedure shows that any input can be rewritten as $\boldsymbol{u} = \sum_i^{\epsilon}\lambda\boldsymbol{v}_i$ such that there are $\epsilon\leq4n$ unique elements in $\boldsymbol{u}$, $\boldsymbol{v}_i\in\{0, 1\}^{4n}$, $0\leq\lambda_i\leq f_{max}$, and $\Sigma_i\lambda_i\leq f_{max}$.
    \textbf{a.} if $\boldsymbol{u=0}$, terminate; else mark the current iteration as~$i$.
    \textbf{b.} Find the smallest entry in $\boldsymbol{u}$, mark its amplitude as $\lambda_i$.
    \textbf{c.} the $j$-th element in $\boldsymbol{v}_i$ is assigned as $v_{ij} = 1$ if and only if $u_j\geq\lambda_i$, otherwise, $v_{ij} = 0$.
    \textbf{d.} $\boldsymbol{u} = \boldsymbol{u} - \lambda_i\boldsymbol{v}_i$, go to step \textbf{a.}
In the procedure, we can see $\boldsymbol{v}_i\in\{0, 1\}^{4n}$ because their components are either $0$ or $1$. There is no repetitive appearance of $\boldsymbol{v}_i$ vectors because otherwise the identical vectors can merge by adding up the corresponding amplitudes. Therefore, $\boldsymbol{Av}_1, \dots, \boldsymbol{Av}_{\epsilon}\in\mathcal{S}$, and $\frac{f_{max}}{\Sigma_i\lambda_i}\boldsymbol{Au}\in\mathcal{H(S)}$ by the definition of the convex hull. Since $\boldsymbol{0}\in\mathcal{H(S)}$, $\boldsymbol{w=Au}$ is on the segment between $\boldsymbol{0}$ and $\frac{f_{max}}{\Sigma_i\lambda_i}\boldsymbol{Au}$, $\boldsymbol{w}\in\mathcal{H(S)}$, which shows $\mathcal{W}\subseteq\mathcal{H(S)}$.
Thus, $\mathcal{H(S)}\equiv\mathcal{W}$. 
\end{proof}

\begin{algorithm}[t]
\small
    \SetKwInOut{Input}{Input}
    \SetKwInOut{Output}{Output}

    \textbf{Function} \textbf{ConstructHull} $(\boldsymbol{A})$\;
    \Input{$\boldsymbol{A}$: configuration matrix of the structure\\}
    \Output{$\mathcal{H(W)}$: convex hull of the feasible wrenches}
    
    \uIf{$\boldsymbol{A}$.columns $\leq$ 1}{
        $\mathcal{S}$ := $\{\boldsymbol{Au}\vert\boldsymbol{u}\in\{0, f_{max}\}^{\boldsymbol{A}\text{.columns}}\}$\\
        \Return \textbf{QuickHull}($\mathcal{S}$)\\
    }
    \Else{
        $\mathcal{H}_1$:= \!\textbf{ConstructHull}(first \!$\lceil\frac{\boldsymbol{A}\text{.cols}}{2}\rceil$\! columns of $\boldsymbol{A}$)\\
        $\mathcal{H}_2$:= \!\textbf{ConstructHull}(last \!$\lfloor\frac{\boldsymbol{A}\text{.cols}}{2}\rfloor$\! columns of $\boldsymbol{A}$)\\
        $\mathcal{V}$ := $\emptyset$\\
        \For{$\boldsymbol{v}_1\in\mathcal{H}_1$.vertices, $\boldsymbol{v}_2\in\mathcal{H}_2$.vertices}{
            $\mathcal{V}\leftarrow\{\boldsymbol{v}_1+\boldsymbol{v}_2\}\cup\mathcal{V}$
        }
        \Return \textbf{QuickHull}($\mathcal{V}$)
    }
    \caption{Divide-and-conquer approach to find the convex hull of the feasible wrenches for a multi-rotor structure}
    \label{Alg:obtainhull}
\end{algorithm}
\setlength{\textfloatsep}{0.5em}
Computing all combinations of the binary inputs, \textit{i.e.,} combinatorial operations of columns vectors in~$\boldsymbol{A}$,  induces exponential time complexity with respect to the number of modules and rotors if not optimized.
Therefore, we present a divide-and-conquer algorithm that constructs $\mathcal{H(S)}$ with an upper bound of exponential time complexity in Algorithm~\ref{Alg:obtainhull}.
If $\boldsymbol{A}$ is a single-column matrix, we calculate the corresponding convex hull, whose vertices are the $\boldsymbol{0}$ vector and the column vector itself, as the base case using the \textbf{QuickHull} algorithm as shown on lines 2 to 4. On lines 5 to 12, if the input configuration matrix has more than $1$ column, we first divide the configuration matrix into two halves, then recursively compute the convex hulls based on the two halved configuration matrices. 
Based on the definition of $\mathcal{S}$, the convex hull for the input configuration matrix is equal to the convex hull of the Minkowski-sum of the convex hulls for the two half configuration matrices, which is also a polytope.
Since the Minkowski-sum of convex polytopes is commutative~\cite{schneider2014convex}, the convex hull of the Minkowski-sum of two convex polytopes is equivalent to the convex hull of the Minkowski-sum of the vertices of each polytope, which is captured on lines 9 to 11 in Algorithm \ref{Alg:obtainhull}. 

We analyze the time complexity of Algorithm \ref{Alg:obtainhull} using Master theorem~\cite{cormen2009introduction}. In each recurrence, the function calls itself with half of the original problem size. Then, it traverses through the vertices found on both hulls and calls \textbf{QuickHull} on the collected vertices to merge the divided problems. The complexity of the merging linearly increases with the number of vertices found on the sub-polytopes. Assume the number of columns is $c$ in the current iteration, the recurrence can be expressed as $T(c) = 2T(\frac{c}{2}) + h(c, d)$, where $d$ represents the number of dimensions of the hull and $h(c, d)$ is a function that is polynomial when $d \leq 3$, and converges to an exponential function of $c$ as $d$ increases according to the authors of the \textbf{QuickHull} algorithm~\cite{barber1996quickhull}. 
Applying the Master theorem, we obtain $T(n) = \Theta(h(n, d))$, which shows our algorithm of constructing the convex hull has a time complexity converging to exponential. When $d = 6$, $h$ is subexponential, \textit{i.e.,} polynomial. In addition, we denote $t = \vert\mathcal{T}\vert$ as the number of elements in $\mathcal{T}$, then determining whether $\mathcal{T}\subseteq\mathcal{W}$ has time complexity of $O(tf)$, where $f$ is the number of faces on $\mathcal{H(S)}$ which is in the order of $O(n^6)$. Thus, the overall complexity of this approach is $\Theta(h(n, d))$.




}

\subsection{Optimization}
\label{sec:opt}
\noindent
Finding $\mathcal{H(W)}$ involves the geometric construction of a convex hull, whose computational complexity suffers from the curse of dimensionality~\cite{HINRICHS2011955}, especially due to the function $h(n, d)$. In fact, since the generators of the zonotope $\mathcal{W}$ are in~$\mathbb{R}^6$, the number of the intermediate vertices, \textit{i.e.,} $\boldsymbol{v}_1 + \boldsymbol{v}_2$ on lines 9 to 11 in Algorithm \ref{Alg:obtainhull} are most likely to end up being the vertices of $\mathcal{H(W)}$~\cite{raynaud_1970}, which is unoptimizable using any technique without losing precision. When the number of elements in $\mathcal{T}$ is small, generating $\mathcal{H(W)}$ then check against the elements in $\mathcal{T}$ may be a waste of computational power.

Inspired by the optimization method used in zonotope containment problems~\cite{DEZA2022101809}, we develop a optimization-based method. The optimization-based approach to determine if a structure of configuration matrix $\boldsymbol{A}$ satisfies a task requirement $\mathcal{T}$ iterates through vectors $\boldsymbol{w}\in\mathcal{T}$ and check if they belong to $\mathcal{W}$. Using the convexity of a zonotope, instead of dealing with the infinite set $\mathcal{W}$, we can compare the maximum magnitude on the boundary point of $\mathcal{W}$ in the direction of $\boldsymbol{w}$. Let $\boldsymbol{w} = \vert w\vert\boldsymbol{\hat w}$ where $\vert w \vert$ and $\boldsymbol{\hat w}$ are the magnitude and unit direction vector of $\boldsymbol{w}$, respectively. By maximizing the wrench in the direction of $\boldsymbol{\hat w}$, \textit{i.e.,} 

\begin{equation}
    \begin{aligned}
    & \underset{\boldsymbol{u}}{\text{maximize}}
    & & \lambda,\\
    & \text{subject to}
    & & 0\preceq\boldsymbol{u}\preceq f_{max},\\
    & & & \boldsymbol{Au} = \lambda\boldsymbol{\hat w},
    \end{aligned}
    \label{eq:convexopt}
\end{equation}
By solving \eqref{eq:convexopt} for all $\boldsymbol{\hat w}$ that is on the unit 6-sphere, we obtain the maximum magnitude of all possible wrenches achievable by the multi-rotor, which bound $\mathcal{H(W)}$. Fig.~\ref{fig:polytopes} shows one such convex hull for the structure shown in Fig. \ref{fig:examplestructure}. A specific representation of the force actuation is presented in \cite{hamandi2021design}. Their approach focuses on the force capabilities, constraining the torque to $\boldsymbol{0}$ when generating the polytope of all feasible forces. In our general representation, we can take such constraints into consideration by adding $\boldsymbol{\tau} = \boldsymbol{0}$ as an additional equality constraint in \eqref{eq:convexopt}.
we obtain the maximum feasible actuation
$\lambda$ in the direction of $\boldsymbol{\hat w}$.  
If $\lambda \geq \vert w\vert$, 
the vehicle can generate the wrench $\boldsymbol{w}$ since it is within its actuation limits.
We repeat the procedure for all elements in $\mathcal{T}$ to determine if $\mathcal{T}\subseteq\mathcal{W}$.

The maximization problem in \eqref{eq:convexopt} finds the wrench vector on the boundary of $\mathcal{W}$ in the direction of $\boldsymbol{\hat w}$. When $\boldsymbol{u=0}$, all rotors of the multi-rotor stop rotating. In this case, the equality constraint $\boldsymbol{Au} = \lambda\boldsymbol{\hat w}$ yields $\lambda=0$, which gives a lower bound $\lambda\geq0$ for all $\boldsymbol{\hat w}$. We apply the norm operator on both sides of the equality constraint, which yields $\Vert\boldsymbol{Au}\Vert = \Vert\lambda\boldsymbol{\hat w}\Vert$. Since $\Vert\boldsymbol{\hat w}\Vert=1$ and $\lambda\geq0$, we obtain $\lambda = \Vert\boldsymbol{Au}\Vert$. Thus, the objective of maximizing $\lambda$ in \eqref{eq:convexopt} is equivalent to maximizing $\lambda^2 = \boldsymbol{u}^\top\!\boldsymbol{A}^\top\!\boldsymbol{Au}$, which shows that it is equivalent to a convex quadratic programming optimization problem. Therefore, we can obtain the solution in time complexity of $O(((4n)^2+4n\cdot4n)^2(4n)^4) =O(n^8)$ with extended Karmarkar's projective algorithm~\cite{ye1989extension,anstreicher1986monotonic} in polynomial time. Determining whether $\mathcal{T}\subseteq\mathcal{W}$ has time complexity of $O(tn^8)$. In comparison to the convex hull based approach, although this approach is less time-efficient when checking $\mathcal{T}\subseteq\mathcal{W}$, it does not require a pre-computation, \textit{i.e.}, generating the convex hull in Algorithm~\ref{Alg:obtainhull}. Therefore, when $\vert\mathcal{T}\vert$ is small, the optimization-based approach provides better time-complexity.

\setlength{\textfloatsep}{1em}
\begin{algorithm}[t!]
\small
    \SetKwInOut{Input}{Input}
    \SetKwInOut{Output}{Output}
    \textbf{Function} \textbf{GenerateConfigSymmetry($\mathcal{D}$, $n_{max}$)}\;
    \Input{$\mathcal{D}$: current structure configuration\\
    $n_{max}$: maximum number of modules added}
    \Output{$\mathcal{S}_1, \dots, \mathcal{S}_{n_{max}}$: all possible configurations, grouped by the number of modules}
    $\mathcal{S}_{0} := \{\mathcal{D}\}$
    \For{$n\in\{0, 1, 2, \dots, n_{max}-1\}$}{
        \For{$\mathcal{D}_i\in \mathcal{C}_n$}{
            $\mathcal{S}_{n+1} := \{\}$\\
            \For{$s\in$ attachable surface on $\mathcal{D}_i$}{
                $s_{sym}$ := the attachable surface on $\mathcal{D}_i$ that is centrosymmetric to $s$\\
                $\mathcal{D}_{temp}$ := $\mathcal{D}_i$ with one more module attached at $s$ and $s_{sym}$\\
                $\mathcal{S}_{n+1}$ = $\{\mathcal{D}_{temp}\}\cup\mathcal{S}_{n+1}$ 
            }
        }
    }
    \caption{Configuration generation method that utilizes the property of symmetry to ensure torque-balance.}
    \label{Alg:heuristic}
\end{algorithm}

\section{Task Adaptation}{
\label{sec:task}
The main challenges in finding a suitable configuration that satisfies a set of task requirements include the uncertain number of modules and the quadratically increasing number of possible configurations as the number of modules increases. Such a problem is different from a pattern formation problem such as in~\cite{Dutta2019Distributed}, or a reconfiguration problem such as in~\cite{yoshida2002self} that are well studied in the multi-agent literature. In our case, the cost of locomotion is ignored and the destination configuration is unknown.

We present two methods to adapt to a task requirement. We can add rotors to a multi-rotor one by one to reconfigure, but defining the position and orientation of each rotor with respect to the new structure during the process of reconfiguration would lead us to a difficult problem. Therefore, we leverage the modularity. In our previous work~\cite{jiawei2021HModQuad}, we defined \emph{torque-balanced modules} that form structures.
Similar to modules, structures satisfy the torque-balance property; so they generate zero torque when all motors have the same input. It is feasible to control a multi-rotor vehicle that is not torque-balanced~\cite{doi:10.1146/annurev-control-042920-012045}. However, the torque-balance property reduces the complexity in its structural configuration. In this paper, we use a module that is torque-balanced and expands its feasible wrenches after connecting with other modules.
\begin{definition}[$T$-Module]
A \emph{$T$-module} is a module with its rotors tilted an angle $\eta_j$ around an axis defined by their arm vector $\boldsymbol{p}_{j}$,
$j=1, \dots, 4$. The angles satisfy $\eta_{1} = \eta_{3} = -\eta_{2} = -\eta_{4}$. Making $\eta = \eta_{1}$, we characterize a $T$-module with a single parameter~$\eta$.
\end{definition}
\noindent
We use homogeneous $T$-modules with $\eta=\frac{\pi}{4}$ to compose structures. 
In Sec.~\ref{sec:polytope}, we provide methods of checking if a vehicle satisfies the required wrenches. We utilize these methods to adapt a structure configuration to a task requirement by reconfiguration. 
\subsection{Adaptation by exhaustive search}
\noindent
We present the exhaustive search-based approach to find the configuration with minimum number of modules given the task requirement $\mathcal{T}$, which evaluates \emph{all} possible structures with up to $n_{max}$ additional modules. Given $\mathcal{T}$ and the initial design $\mathcal{D}$, we dock the modules one by one to the attachable surfaces of the possible current design to ensure the structure is connected, and check if the structure 
\textit{i)} is torque-balanced, and \textit{ii)} satisfies $\mathcal{T}$. The process constrains the relative position of the new module to the structure during the docking so that the new module aligns one of its surfaces with a module in the structure. 

To ensure all possible configurations are evaluated, every time a new module is added, the algorithm iterates through all designs from the previous step as possible current designs, and evaluates if connecting a new module to any one of them makes it a structure that satisfies \textit{i)} and \textit{ii)} using the methods introduced in Sec.~\ref{sec:polytope}. Therefore, this approach guarantees minimum number of modules added to the initial design. 

The exhaustive search evaluates all possible configurations. However, as modules are added to the configuration, the COM of the structure changes, and the potential asymmetry may degrade the control quality. To avoid these problems, we take a heuristic approach in practice.

\subsection{Heuristic task adaptation}{

\noindent
When adding modules to a structure, certain patterns apply to ensure that a structure is torque-balanced. Moreover, adding modules to a structure may cause the COM to shift with respect to the structure frame. Some aerial tasks do not allow the COM to shift. For instance, a structure is equipped with a special mechanism to carry a payload, and we want to increase the maximum weight of the payload the structure can carry. We have to make sure the new configuration does not change the COM. Otherwise, the structure may lose control when transporting the payload. Therefore, we introduce the centrosymmetry pattern that can help reduce the search space and makes sure the COM stays in place during the reconfiguration. 

When generating all possible designs, instead of adding modules one by one and iterating through all attachable surfaces of all possible designs from the previous step, we call \textbf{GenerateConfigSymmetry} in Algorithm \ref{Alg:heuristic} that applies the heuristics of centrosymmetry to the exhaustive search. On line 6 of Algorithm \ref{Alg:heuristic}, in addition to adding the module at an attachable surface $s$ on the structure, we also find the surface $s_{sym}$ that is centrosymmetric to $s$ with respect to the COM of the existing configuration, and add another module at $s_{sym}$. Since all modules are homogeneous, the COM of the new structure remains the same. Note that centrosymmetry is not a necessary, but a sufficient condition for the assembled structure to be torque-balanced and to have a unshifted COM. It is possible that a non-symmetric structure is torque-balanced, or reconfiguring a structure without following the centrosymmetry rule causes no shift on the COM. 

}
}

\begin{figure}
\centering
\vspace{0.5em}
    \includegraphics[trim=0cm 5cm 0cm 1cm, clip, width=0.8\linewidth]{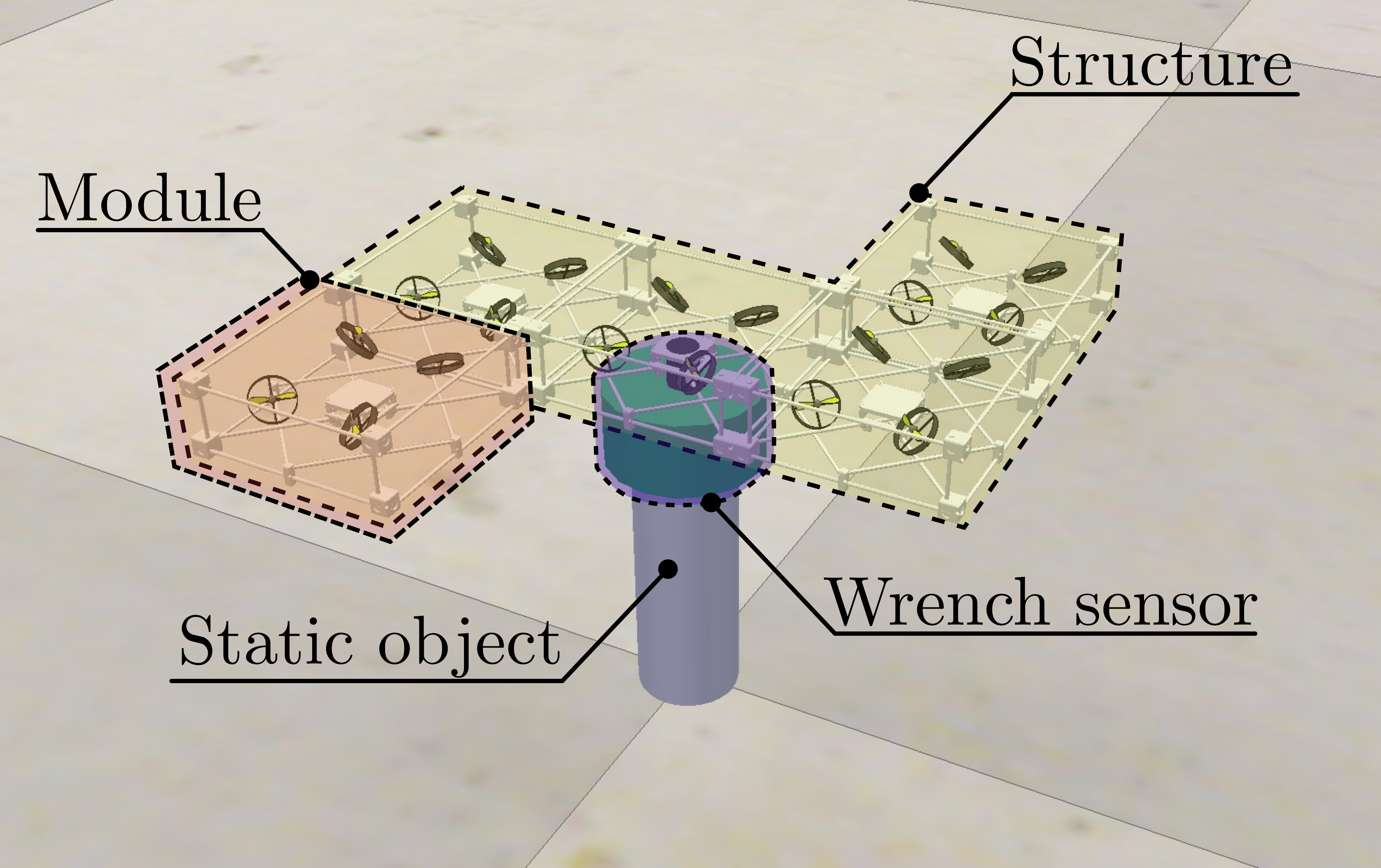}
    \caption{Setup for the simulation in CoppeliaSim. 
    }
    \label{fig:simexp}
\end{figure}

\section{Evaluation}
\setlength{\textfloatsep}{2em}
\label{sec:eval}
We implement the algorithms introduced in Sec.~\ref{sec:polytope} and~\ref{sec:task}\footnote{The source code can be found on \url{https://github.com/Jarvis-X/HModQuad-sim}}. We provide a task requirement for both the exhaustive search-based and the heuristic approaches to find structure configurations that satisfy the requirement. The task requirement is composed of randomly generated wrenches with an offset towards more force along the positive $z$-axis, $f_z$, to emulate the actuation to compensate the gravity. 
For both algorithms, we start with an initial structure of one module and set the maximum number of modules $n_{max}=7$. We assume the maximum thrust force for each rotor $f_{max} = 1$, and the side length $l=0.4$ for each module.

After the algorithms return structure configurations that satisfy the requirement, we construct the structures in CoppeliaSim~\cite{coppeliaSim} environment. Since we want to show that the structure is able to generate the wrenches in the task requirement, we use a ``force sensor'' component to rigidly connect the COM of the structure to a static wall. All wrenches generated by the structure are transferred through the force sensor to the wall, which measures the wrench generated by the structure without causing any motion. Fig.~\ref{fig:simexp} shows an example of such setup in CoppeliaSim.
\begin{figure}
\vspace{-1em}
\centering
    \subfloat[Exhaustive Search\label{fig:randomexh_config}]{\includegraphics[width=0.32\linewidth]{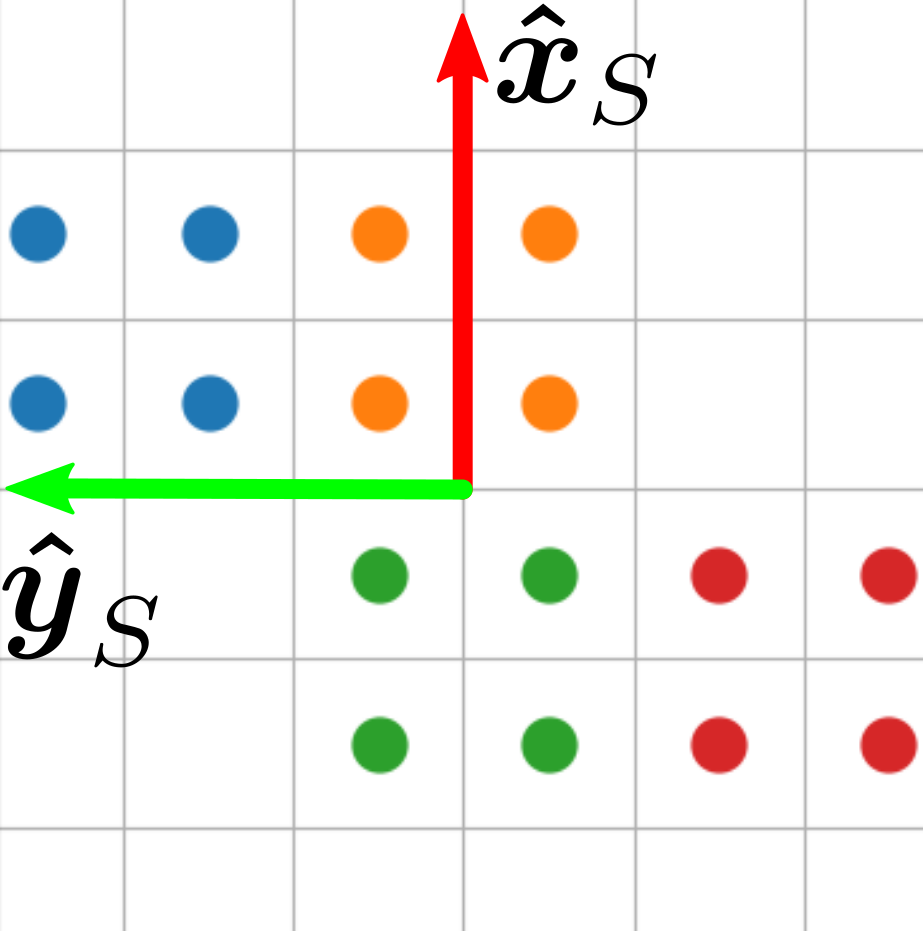}}
    \hspace{0.05\linewidth}
    \subfloat[Heuristic Search\label{fig:randomh_config}]{\includegraphics[width=0.32\linewidth]{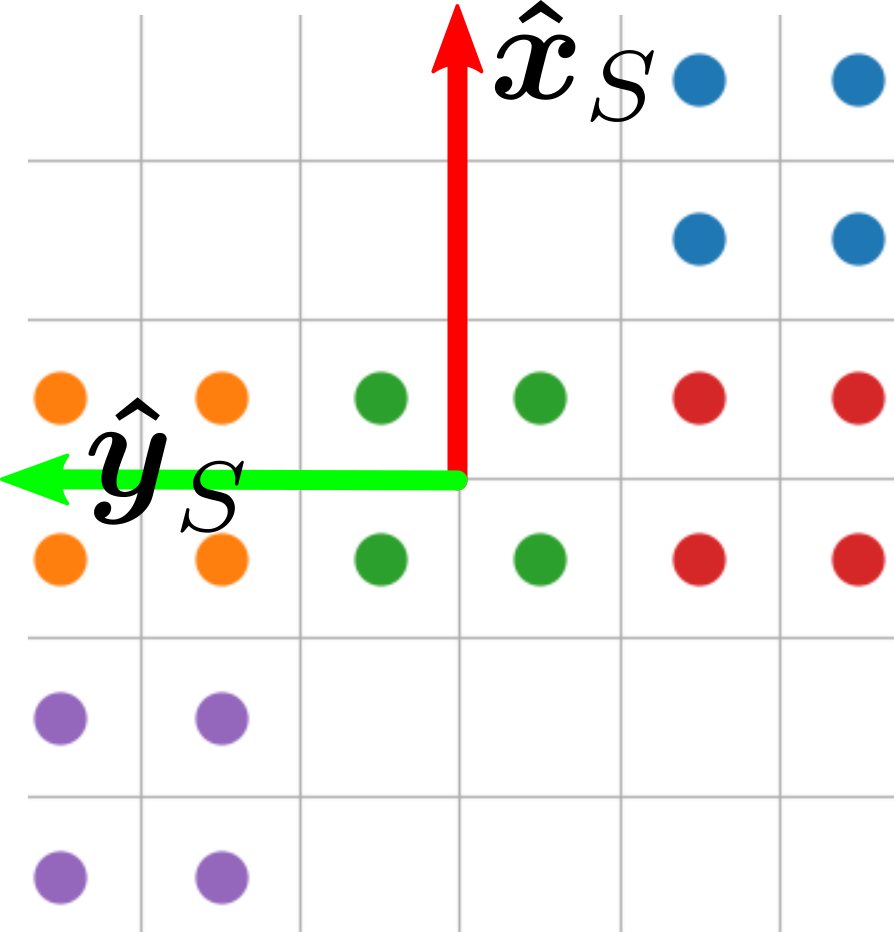}}
    \caption{Configurations found by our methods.}
    \label{fig:configurations}
    \vspace{-1em}
\end{figure}
We randomly generate a task requirement $\mathcal{T}$ that is composed of $80$ wrenches. Each element of the wrenches is a random value uniformly chosen on $(-0.5, 0.5)$, and we multiply $30$ on $f_z$ to emphasize the gravity. $\mathcal{T}$ is the input to both exhaustive and heuristic search algorithms. The exhaustive search-based approach evaluates 9 configurations and returns a configuration of 4 modules as shown in Fig. \ref{fig:randomexh_config}. The heuristic approach evaluates 4 configurations and returns a configuration of 5 modules as shown in Fig. \ref{fig:randomh_config}. 

\begin{figure}[t!]
\centering
{
    \vspace{0em}
    \subfloat[The rotor input of the configuration found by the \emph{exhaustive search}.\label{fig:random1exh}]{%
       \includegraphics[trim=0cm 0cm 0.5cm 0cm, clip, width=0.48\linewidth]{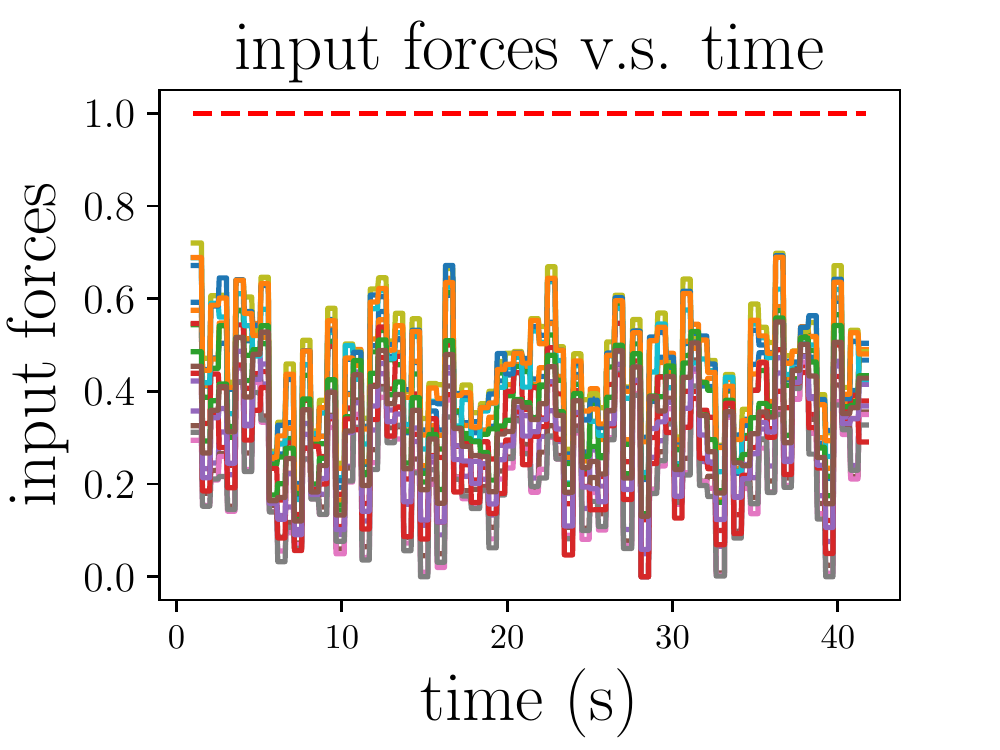}}
    \hspace{0.02\linewidth}
    \subfloat[The rotor input of the configuration found by the \emph{heuristic search}.\label{fig:random1h}]{%
        \includegraphics[trim=0cm 0cm 0.5cm 0cm, clip, width=0.48\linewidth]{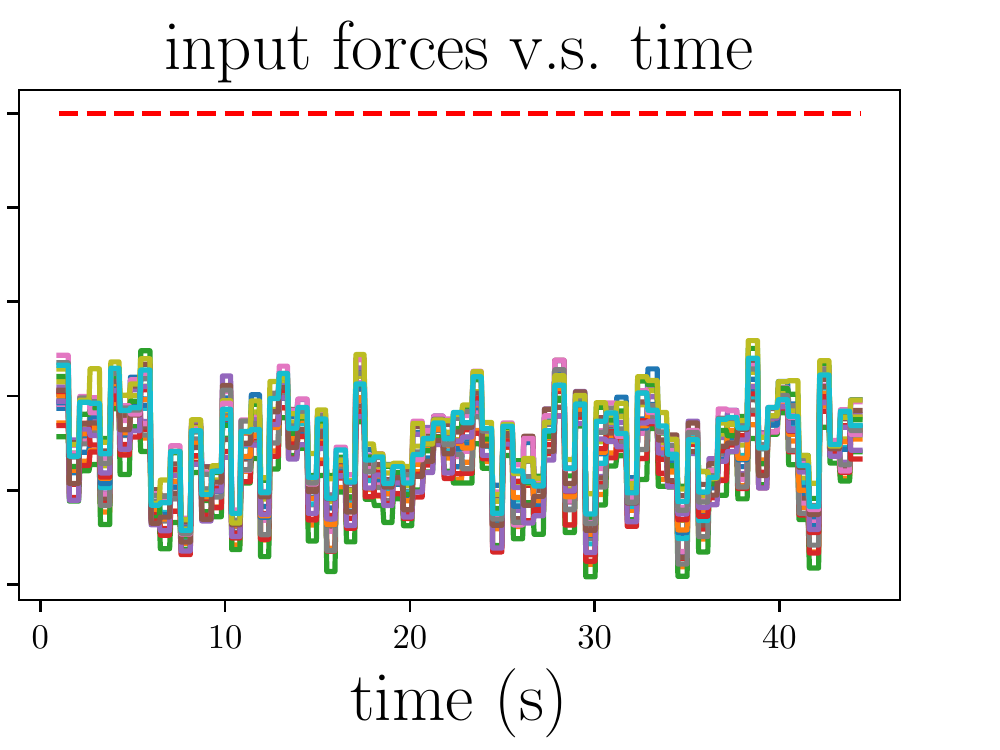}}
    \caption{The structure configurations found by both algorithms satisfy the task requirement without input saturation. The dot line represents $f_{max}$, and the solid lines represent the input forces of the rotors.}
    \label{fig:random1}
    \vspace{-1em}
    }
\end{figure}

We actuate the constructed structures. In our previous work~\cite{jiawei2021HModQuad}, we present control strategies for a fully-actuated structure. Here, since the structures are rigidly connected by a force sensor to a static wall, their motion properties are ignored. Therefore, we apply a simplified approach. For a configuration, we calculate the desired input $\boldsymbol{u^\star = A}^\dagger\boldsymbol{w}$ for each $\boldsymbol{w}\in\mathcal{T}$ every 500 milliseconds. Since each rotor can produce thrust force between $0$ and $1$, we truncate the input vector $\boldsymbol{u^\star}$ outside the range of $\left[0, 1\right]$. The truncated input vector $0\preceq\boldsymbol{u}\preceq1$ is then taken as the effective thrust forces the rotors of a structure generate. As shown in Fig. \ref{fig:random1exh} and \ref{fig:random1h}, the configurations provided by both algorithms can generate all required wrenches without input saturation.

The simulations for the random task requirement show that the structure configurations found by both algorithms can generate the required wrenches without saturating the motors as shown in Fig.~\ref{fig:random1}, which validates our methods. 
Fig.~\ref{fig:configurations} illustrates the configurations, where every four circles of the same color represent rotors of one module, and the green circles belong to the initial module. The exhaustive search-based algorithm outputs the configuration in Fig.~\ref{fig:randomexh_config}, which is centrosymmetric but with shifted COM from the initial configuration. It provides the optimal solution to the task requirement as for the less number of modules in comparison to the configuration found by the heuristic method in Fig.~\ref{fig:randomh_config}. 

\section{Conclusion and Future Work}{
\label{sec:conclusion}
}
In this paper, we have presented a comprehensive framework for analyzing the actuation capabilities of modular multi-rotor vehicles and determining their ability to satisfy task requirements. Our contributions include a formal definition of aerial task requirements, a geometric model of the feasible wrenches of a multi-rotor vehicle with uni-directional rotors, and a convex hull abstraction of the actuation capabilities of a multi-rotor vehicle. We have also introduced an optimization-based algorithm to identify whether a configuration satisfies a set of task requirements, as well as a search-based method for expanding the capabilities of modular multi-rotor vehicles to satisfy task requirements.

Our methods have been demonstrated using the H-ModQuad platform, which features tilted rotors to increase the number of ADOF. However, our abstractions and algorithms are applicable to any modular multi-rotor vehicle. We have shown that our approach is effective in finding the optimal configuration of homogeneous modules that satisfy task requirements.

For future work, we plan to extend our methods to heterogeneous modules and test our structures with real robots. We would also like to explore the possibility of incorporating the dynamics of the structures into the task requirements to enable motion during evaluation. By doing so, we can further improve the adaptability and versatility of modular multi-rotor vehicles in various applications.


\section*{Acknowledgement}{
The authors thank Diego S. D'Antionio at Lehigh University for the module design and renderings.
}
\bibliographystyle{IEEEtran}
\bibliography{ref}

\begin{thebibliography}{10}
\providecommand{\url}[1]{#1}
\csname url@samestyle\endcsname
\providecommand{\newblock}{\relax}
\providecommand{\bibinfo}[2]{#2}
\providecommand{\BIBentrySTDinterwordspacing}{\spaceskip=0pt\relax}
\providecommand{\BIBentryALTinterwordstretchfactor}{4}
\providecommand{\BIBentryALTinterwordspacing}{\spaceskip=\fontdimen2\font plus
\BIBentryALTinterwordstretchfactor\fontdimen3\font minus
  \fontdimen4\font\relax}
\providecommand{\BIBforeignlanguage}[2]{{%
\expandafter\ifx\csname l@#1\endcsname\relax
\typeout{** WARNING: IEEEtran.bst: No hyphenation pattern has been}%
\typeout{** loaded for the language `#1'. Using the pattern for}%
\typeout{** the default language instead.}%
\else
\language=\csname l@#1\endcsname
\fi
#2}}
\providecommand{\BIBdecl}{\relax}
\BIBdecl

\bibitem{1302409}
S.~{Bouabdallah}, P.~{Murrieri}, and R.~{Siegwart}, ``Design and control of an
  indoor micro quadrotor,'' in \emph{IEEE International Conference on Robotics
  and Automation, 2004. Proceedings. ICRA '04. 2004}, vol.~5, 2004, pp.
  4393--4398 Vol.5.

\bibitem{7139759}
S.~{Rajappa}, M.~{Ryll}, H.~H. {Bülthoff}, and A.~{Franchi}, ``Modeling,
  control and design optimization for a fully-actuated hexarotor aerial vehicle
  with tilted propellers,'' in \emph{2015 IEEE International Conference on
  Robotics and Automation (ICRA)}, 2015, pp. 4006--4013.

\bibitem{alaimo2013mathematical}
A.~Alaimo, V.~Artale, C.~Milazzo, A.~Ricciardello, and L.~Trefiletti,
  ``Mathematical modeling and control of a hexacopter,'' in \emph{2013
  International Conference on Unmanned Aircraft Systems (ICUAS)}.\hskip 1em
  plus 0.5em minus 0.4em\relax IEEE, 2013, pp. 1043--1050.

\bibitem{beniak2016control}
R.~Beniak and O.~Gudzenko, ``Control methods design for a model of asymmetrical
  quadrocopter,'' \emph{Journal of Automation Mobile Robotics and Intelligent
  Systems}, vol.~10, 2016.

\bibitem{Oung2011TheArray}
\BIBentryALTinterwordspacing
R.~Oung and R.~D'Andrea, ``{The distributed flight array},''
  \emph{Mechatronics}, vol.~21, no.~6, pp. 908--917, 2011. [Online]. Available:
  \url{http://dx.doi.org/10.1016/j.mechatronics.2010.08.003}
\BIBentrySTDinterwordspacing

\bibitem{8461014}
D.~{Saldaña}, B.~{Gabrich}, G.~{Li}, M.~{Yim}, and V.~{Kumar}, ``Modquad: The
  flying modular structure that self-assembles in midair,'' in \emph{2018 IEEE
  International Conference on Robotics and Automation (ICRA)}, 2018, pp.
  691--698.

\bibitem{8258850}
M.~Zhao, T.~Anzai, F.~Shi, X.~Chen, K.~Okada, and M.~Inaba, ``Design, modeling,
  and control of an aerial robot dragon: A dual-rotor-embedded multilink robot
  with the ability of multi-degree-of-freedom aerial transformation,''
  \emph{IEEE Robotics and Automation Letters}, vol.~3, no.~2, pp. 1176--1183,
  2018.

\bibitem{jiawei2021HModQuad}
J.~Xu, D.~S. D’Antonio, and D.~Saldaña, ``H-modquad: Modular multi-rotors
  with 4, 5, and 6 controllable dof,'' in \emph{2021 IEEE International
  Conference on Robotics and Automation (ICRA)}, 2021, pp. 190--196.

\bibitem{9636086}
B.~Gabrich, D.~Saldaña, and M.~Yim, ``Finding structure configurations for
  flying modular robots,'' in \emph{2021 IEEE/RSJ International Conference on
  Intelligent Robots and Systems (IROS)}, 2021, pp. 6970--6976.

\bibitem{7759271}
M.~Ryll, D.~Bicego, and A.~Franchi, ``Modeling and control of fast-hex: A
  fully-actuated by synchronized-tilting hexarotor,'' in \emph{2016 IEEE/RSJ
  International Conference on Intelligent Robots and Systems (IROS)}, 2016, pp.
  1689--1694.

\bibitem{hamandi2021design}
M.~Hamandi, F.~Usai, Q.~Sabl{\'e}, N.~Staub, M.~Tognon, and A.~Franchi,
  ``Design of multirotor aerial vehicles: a taxonomy based on input
  allocation,'' 2021.

\bibitem{6161298}
R.~Naldi, F.~Forte, and L.~Marconi, ``A class of modular aerial robots,'' in
  \emph{2011 50th IEEE Conference on Decision and Control and European Control
  Conference}, 2011, pp. 3584--3589.

\bibitem{9438375}
J.~Strawson, P.~Cao, T.~Bewley, and F.~Kuester, ``Rotor orientation
  optimization for direct 6 degree of freedom control of multirotors,'' in
  \emph{2021 IEEE Aerospace Conference (50100)}, 2021, pp. 1--12.

\bibitem{6935598}
O.~Magnussen, G.~Hovland, and M.~Ottestad, ``Multicopter uav design
  optimization,'' in \emph{2014 IEEE/ASME 10th International Conference on
  Mechatronic and Embedded Systems and Applications (MESA)}, 2014, pp. 1--6.

\bibitem{halperin2020space}
D.~Halperin, M.~v. Kreveld, G.~Miglioli-Levy, and M.~Sharir, ``Space-aware
  reconfiguration,'' in \emph{International Workshop on the Algorithmic
  Foundations of Robotics}.\hskip 1em plus 0.5em minus 0.4em\relax Springer,
  2020, pp. 37--53.

\bibitem{9172614}
K.~Garanger, J.~Epps, and E.~Feron, ``Modeling and experimental validation of a
  fractal tetrahedron uas assembly,'' in \emph{2020 IEEE Aerospace Conference},
  2020, pp. 1--11.

\bibitem{Mellinger2012CooperativeQuadrotors}
D.~Mellinger, M.~Shomin, N.~Michael, and V.~Kumar, ``{Cooperative grasping and
  transport using multiple quadrotors},'' \emph{Springer Tracts in Advanced
  Robotics}, vol. 83 STAR, pp. 545--558, 2012.

\bibitem{9364354}
D.~S. D’Antonio, G.~A. Cardona, and D.~Saldaña, ``The catenary robot: Design
  and control of a cable propelled by two quadrotors,'' \emph{IEEE Robotics and
  Automation Letters}, vol.~6, no.~2, pp. 3857--3863, 2021.

\bibitem{duffy2015lift}
M.~J. Duffy and T.~C. Samaritano, ``The lift! project--modular, electric
  vertical lift system with ground power tether,'' in \emph{33rd AIAA Applied
  Aerodynamics Conference}, 2015, p. 3013.

\bibitem{modquadgripper}
B.~Gabrich, D.~Saldaña, V.~Kumar, and M.~Yim, ``A flying gripper based on
  cuboid modular robots,'' 05 2018.

\bibitem{6225129}
M.~{Ryll}, H.~H. {Bülthoff}, and P.~R. {Giordano}, ``Modeling and control of a
  quadrotor uav with tilting propellers,'' in \emph{2012 IEEE International
  Conference on Robotics and Automation}, 2012, pp. 4606--4613.

\bibitem{5569026}
N.~{Michael}, D.~{Mellinger}, Q.~{Lindsey}, and V.~{Kumar}, ``The grasp
  multiple micro-uav testbed,'' \emph{IEEE Robotics Automation Magazine},
  vol.~17, no.~3, pp. 56--65, 2010.

\bibitem{boyd2004convex}
S.~Boyd, S.~P. Boyd, and L.~Vandenberghe, \emph{Convex optimization}.\hskip 1em
  plus 0.5em minus 0.4em\relax Cambridge U. press, 2004.

\bibitem{borrelli2017predictive}
F.~Borrelli, A.~Bemporad, and M.~Morari, \emph{Predictive control for linear
  and hybrid systems}.\hskip 1em plus 0.5em minus 0.4em\relax Cambridge
  University Press, 2017.

\bibitem{fukuda2004zonotope}
K.~Fukuda, ``From the zonotope construction to the minkowski addition of convex
  polytopes,'' \emph{Journal of Symbolic Computation}, vol.~38, no.~4, pp.
  1261--1272, 2004.

\bibitem{kuhn1998rigorously}
W.~K{\"u}hn, ``Rigorously computed orbits of dynamical systems without the
  wrapping effect,'' \emph{Computing}, vol.~61, no.~1, pp. 47--67, 1998.

\bibitem{coxeter1973regular}
H.~S.~M. Coxeter, \emph{Regular polytopes}.\hskip 1em plus 0.5em minus
  0.4em\relax Courier Corporation, 1973.

\bibitem{beck2007computing}
M.~Beck and S.~Robins, \emph{Computing the continuous discretely}.\hskip 1em
  plus 0.5em minus 0.4em\relax Springer, 2007, vol.~61.

\bibitem{grunbaum1967convex}
B.~Gr{\"u}nbaum, V.~Klee, M.~A. Perles, and G.~C. Shephard, \emph{Convex
  polytopes}.\hskip 1em plus 0.5em minus 0.4em\relax Springer, 1967, vol.~16.

\bibitem{barber1996quickhull}
C.~B. Barber, D.~P. Dobkin, and H.~Huhdanpaa, ``The quickhull algorithm for
  convex hulls,'' \emph{ACM Transactions on Mathematical Software (TOMS)},
  vol.~22, no.~4, pp. 469--483, 1996.

\bibitem{schneider2014convex}
R.~Schneider, \emph{Convex bodies: the Brunn--Minkowski theory}.\hskip 1em plus
  0.5em minus 0.4em\relax Cambridge U. press, 2014, no. 151.

\bibitem{cormen2009introduction}
T.~H. Cormen, C.~E. Leiserson, R.~L. Rivest, and C.~Stein, \emph{Intro. to
  algorithms}.\hskip 1em plus 0.5em minus 0.4em\relax MIT press, 2009.

\bibitem{HINRICHS2011955}
\BIBentryALTinterwordspacing
A.~Hinrichs, E.~Novak, and H.~Woźniakowski, ``The curse of dimensionality for
  the class of monotone functions and for the class of convex functions,''
  \emph{Journal of Approximation Theory}, vol. 163, no.~8, pp. 955--965, 2011.
  [Online]. Available:
  \url{https://www.sciencedirect.com/science/article/pii/S0021904511000311}
\BIBentrySTDinterwordspacing

\bibitem{raynaud_1970}
H.~Raynaud, ``Sur l'enveloppe convexe des nuages de points aleatoires dans rn.
  i,'' \emph{Journal of Applied Probability}, vol.~7, no.~1, p. 35–48, 1970.

\bibitem{DEZA2022101809}
\BIBentryALTinterwordspacing
A.~Deza and L.~Pournin, ``A linear optimization oracle for zonotope
  computation,'' \emph{Computational Geometry}, vol. 100, p. 101809, 2022.
  [Online]. Available:
  \url{https://www.sciencedirect.com/science/article/pii/S0925772121000651}
\BIBentrySTDinterwordspacing

\bibitem{ye1989extension}
Y.~Ye and E.~Tse, ``An extension of karmarkar's projective algorithm for convex
  quadratic programming,'' \emph{Mathematical programming}, vol.~44, no.~1, pp.
  157--179, 1989.

\bibitem{anstreicher1986monotonic}
K.~M. Anstreicher, ``A monotonic projective algorithm for fractional linear
  programming,'' \emph{Algorithmica}, vol.~1, no.~1, pp. 483--498, 1986.

\bibitem{Dutta2019Distributed}
A.~Dutta, P.~Dasgupta, and C.~Nelson, ``\BIBforeignlanguage{English
  (US)}{Distributed configuration formation with modular robots using
  (sub)graph isomorphism-based approach},'' \emph{\BIBforeignlanguage{English
  (US)}{Autonomous Robots}}, vol.~43, no.~4, pp. 837--857, Apr. 2019, publisher
  Copyright: {\textcopyright} 2018, Springer Science+Business Media, LLC, part
  of Springer Nature.

\bibitem{yoshida2002self}
E.~Yoshida, S.~Murata, A.~Kamimura, K.~Tomita, H.~Kurokawa, and S.~Kokaji, ``A
  self-reconfigurable modular robot: Reconfiguration planning and
  experiments,'' \emph{The International Journal of Robotics Research},
  vol.~21, no. 10-11, pp. 903--915, 2002.

\bibitem{doi:10.1146/annurev-control-042920-012045}
\BIBentryALTinterwordspacing
M.~W. Mueller, S.~J. Lee, and R.~D'Andrea, ``Design and control of drones,''
  \emph{Annual Review of Control, Robotics, and Autonomous Systems}, vol.~5,
  no.~1, pp. 161--177, 2022. [Online]. Available:
  \url{https://doi.org/10.1146/annurev-control-042920-012045}
\BIBentrySTDinterwordspacing

\bibitem{coppeliaSim}
E.~Rohmer, S.~P.~N. Singh, and M.~Freese, ``Coppeliasim (formerly v-rep): a
  versatile and scalable robot simulation framework,'' in \emph{Proc. of The
  International Conference on Intelligent Robots and Systems (IROS)}, 2013,
  www.coppeliarobotics.com.

\end{thebibliography}

\end{document}